\documentclass[letterpaper, 10 pt, conference]{ieeeconf}

\usepackage{cite}
\usepackage{amsmath,amssymb,amsfonts}
\usepackage{graphicx}
\usepackage[hidelinks]{hyperref} 
\usepackage{textcomp}
\usepackage{tabularx}
\usepackage{amsmath}
\usepackage{balance}
\usepackage{multirow}
\usepackage{dblfloatfix}
\usepackage{siunitx}
\usepackage{booktabs}
\usepackage{longtable}
\usepackage{cite}
\usepackage{color,colortbl}
\usepackage{algpseudocode}
\usepackage{algorithm}
\definecolor{Gray2}{gray}{0.9}
\definecolor{Gray}{gray}{0.7}

\usepackage{xcolor}

\IEEEoverridecommandlockouts 

\begin{document}

\title{\LARGE \bf SwarmPath: Drone Swarm Navigation through Cluttered Environments Leveraging Artificial Potential Field and Impedance Control\\
}

\author{Roohan Ahmed Khan\textsuperscript{*}, Malaika Zafar\textsuperscript{*}, Amber Batool\textsuperscript{*}, Aleksey Fedoseev, and Dzmitry Tsetserukou %
\thanks{The authors are with the Intelligent Space Robotics Laboratory, Center for Digital Engineering, Skolkovo Institute of Science and Technology, Moscow, Russia. 
\tt \{roohan.khan, malaika.zafar, amber.batool, aleksey.fedoseev, d.tsetserukou\}@skoltech.ru}
\thanks{\textsuperscript{*}These authors contributed equally to this work.}
}

\maketitle

\begin{abstract}
In the area of multi-drone systems, navigating through dynamic environments from start to goal while providing collision-free trajectory and efficient path planning is a significant challenge. To solve this problem, we propose a novel SwarmPath technology that involves the integration of Artificial Potential Field (APF) with Impedance Controller. The proposed approach provides a solution based on collision free leader-follower behaviour where drones are able to adapt themselves to the environment. Moreover, the leader is virtual while drones are physical followers leveraging APF path planning approach to find the smallest possible path to the target. Simultaneously, the drones dynamically adjust impedance links, allowing themselves to create virtual links with obstacles to avoid them. As compared to conventional APF, the proposed SwarmPath system not only provides smooth collision-avoidance but also enable agents to efficiently pass through narrow passages by reducing the total travel time by $30\%$ while ensuring safety in terms of drones connectivity. Lastly, the results also illustrate that the discrepancies between simulated and real environment, exhibit an average absolute percentage error (APE) of $6\%$ of drone trajectories. This underscores the reliability of our solution in real-world scenarios. 
\end{abstract}


\vspace{1em}
\textbf{Video:} 
\href{https://youtu.be/k8Nf_vPZf7U}{https://youtu.be/k8Nf\textunderscore vPZf7U}

\begin{figure}[htbp]
\centering
\vspace{-0.2cm}
\includegraphics[width=0.95\linewidth]{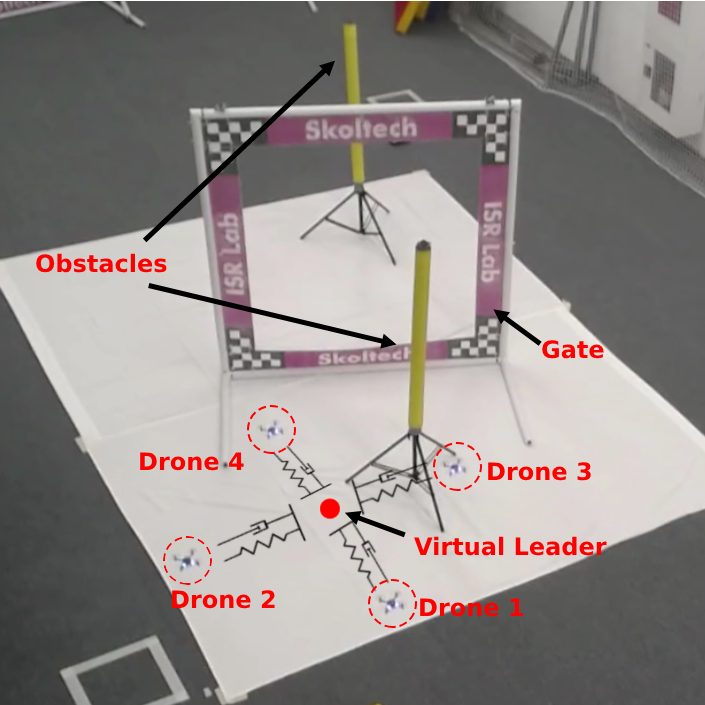} 
\caption{SwarmPath technology performing gate navigation and obstacle avoidance on a complex racing terrain by adaptive impedance link topology where it shows the overall setup of swarm, obstacles and gate position.}
\label{exp_setup_1}
\vspace{-0.2cm}

\end{figure}

\section{Introduction}
Despite the benefits of using swarms of drones in multiple applications, many UAVs currently lack the necessary adaptable programming tools. These tools are required for effective navigation in complex environments. Therefore, many path-planning techniques have been developed in past decades to provide effective   navigation and coordination among drones \cite{AGGARWAL2020270}. 

Xu et al\cite{DRL}. developed a UAV path-planning method using deep reinforcement learning in AirSim with the training policy of path planning for drone swarms but system falls short in adapting to specific use cases.

Another system was introduced by Kaufmann et al. \cite{Drone_Racing} that combines DRL in simulation with data collected in the physical world but drones cannot adjust themselves for missions, including multi-agent system management. 

Kalinov at al. \cite{UAV-UGV} proposed an impedance control system which ensures that the drone lands softly with minimal initial contact force by keeping a smooth connection throughout the landing process.

The SwarmHawk technology was introduced by Gupta et al. \cite{swarmhawk} based on APF for robust landing of swarm of drones on a moving platform, demonstrating sufficient accuracy in homogeneous and leader-follower formations.

Tsykunov et al. \cite{SwarmTouch} proposed guiding the swarms of nano-drones through human-swarm communication, combining impedance control with vibrotactile feedback interface. The proposed system relies on human cognitive ability to guide a swarm of drones in a cluttered environment while autonomously adjusting the swarm formation through virtual impedance links. A similar concept was later utilized by Fedoseev et al. \cite{Dandelion} exploring the topology of virtual impedance links. The research revealed that star topology allows the drones in a swarm to achieve a higher accuracy of path following. However, such experiments rely solely on human-swarm communication, which is not always applicable in real-world scenarios.

 Batinovic et al. \cite{APF_LiDAR} applied APF to tackle path planning problem for aerial robots in unknown environment, prioritizing safe trajectory execution while avoiding complex obstacles with a help of LiDAR sensor. A novel distributed control algorithm that combines the APF method in the virtual leader formation scheme and switching communication network is proposed by Yu et al \cite{Yu_2023}. Mishra et al. \cite{Mishra_2023} proposed a multi-agent DRL-based architecture with graph convolution called GALOPP, which incorporates the limited sensor field-of-view, communication, and localization constraints of the swarm. Moreover, the study of Barclay et al. \cite{Barclay2024HumanguidedSI} presents the human supervisors to tune the level of swarm control by applying VR environment integrated with impedance-control influence mechanism. Experimental results show the completion of tasks by drones while maintaining innate formation characteristics. 

 Luis et al. \cite{inproceedings_vsd} proposed a Virtual Spring-Damper (VSD) approach for controlling UAV swarm formations and avoiding collisions through decentralized force-based mechanism. Their work primarily addresses local interactions and but does not cater for global path planning strategy. 

This paper presents a novel system for agile and safe path following by a swarm of drones in cluttered environments. Our contributions are as follows:
1) \textit{SwarmPath approach}: the development of a dual path planning approach that combines the APF algorithm \cite{APF_article}, \cite{srivastava2024reformulation} for global path planning with impedance links, featuring a novel adaptive topology for local path correction close to obstacles.
2) \textit{Comparative Analysis}:  we explore the limitations of the proposed approach and conduct a comparative analysis with the baseline APF algorithm in simulation with gates and convex obstacles.
3) \textit{Real-World Experiment}: we conduct an experiment in a real-world environment utilizing a swarm of Crazyflie 2.1 drones, evaluating the effectiveness of impedance topology in the context of actual drone dynamics.

\section{Methodology}

Drones and virtual leader points are initialized with states and impedance control parameters. The virtual leader creates a path using APF path planning as global while leaving the interaction of connected drones with obstacles to local path planner. The local path planner keeps the swarm of drones collision-free so that when any drone observes an obstacle, the drone breaks away from the APF trajectory and deflects from the obstacle by creating a virtual link with the respective obstacle. Once the obstacle is avoided, the drone reconnects with the APF's trajectory. This technique helps in smooth implementation of APF trajectory globally while also avoiding all the drones from collision in locally defined regions.

\section{SwarmPath Technology}
\begin{figure}[htbp]
\centering
\includegraphics[width=0.96\linewidth]{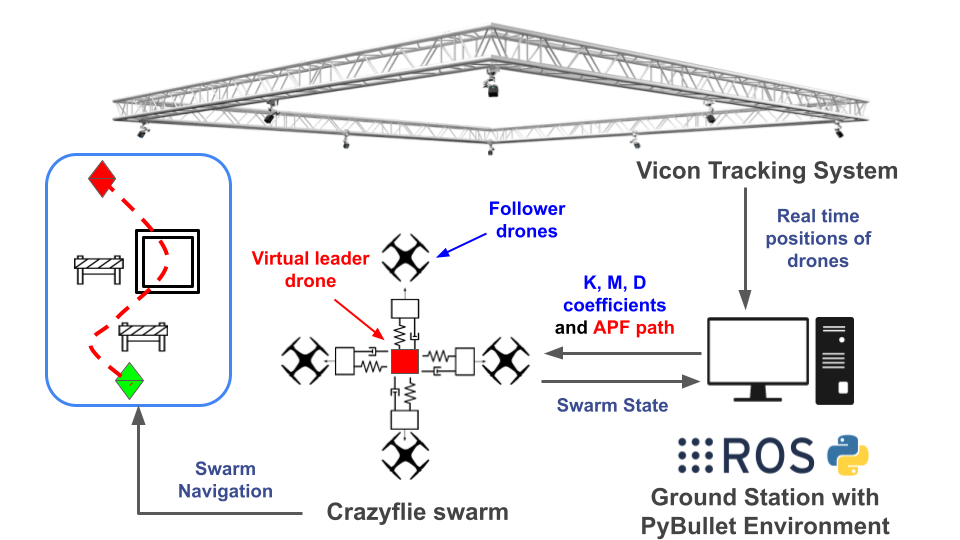} 
\caption{SwarmPath utilizing a swarm of Crazyflie drones, VICON for drone motion tracking, ROS2 for publishing APF path for swarm centroid and impedance parameters for the follower drones. Moreover, PyBullet Environment was used to perform drone simulation.}
\label{exp_setup}
\vspace{-0.3cm}

\end{figure}

The pipeline of the adaptive swarm path planning is shown in Fig. \ref{exp_setup}. This section shows the development of path planning algorithm using APF and impedance control.

\subsection{Impedance Control}
\label{section:impedance}
APF is designed for single-agent path planning, therefore, the trajectory generated by the virtual leader via APF manages the trajectory of all other drones. Since there arises challenges of interference and collision as APF generates force fields around each drone, and also synchronization issues when each drone will generate its own path. To address this issue, we combine APF with impedance controller \cite{Impedance} to enhance swarm navigation. 

\subsubsection{{Leader-Drone Impedance Control}
\label{section:Impedance}} In this model, each drone's position is coupled with APF trajectory, which serves as the leading trajectory, through mass-spring-damper system. This forms the impedance links of the physical drones with the virtual leader. These impedance links prevent collisions among the drones, providing a smooth and efficient trajectory. The links among the drones are established using a second-order differential equation of mass-spring-damper given as:

\begin{equation}
m\Delta\ddot{x} + d\Delta\dot{x} + k\Delta x = F_{\text{ext}}(t)
\label{eq:dynamic_equation}
\end{equation}
where $\Delta x$ is the difference of current and desired drone position and $F_{\text{ext}}(t)$ is the virtual external force applied as an input from leader. $m$ is the virtual mass of a link, $d$ is the damping coefficient of the virtual damper, and $k$ is the virtual spring constant. 

\subsubsection{{Obstacle-Drone Impedance Control}
\label{section:Impedance}}
\begin{figure}[h!]
\vspace{-1.2cm}

\centering
\includegraphics[width=0.98\linewidth]{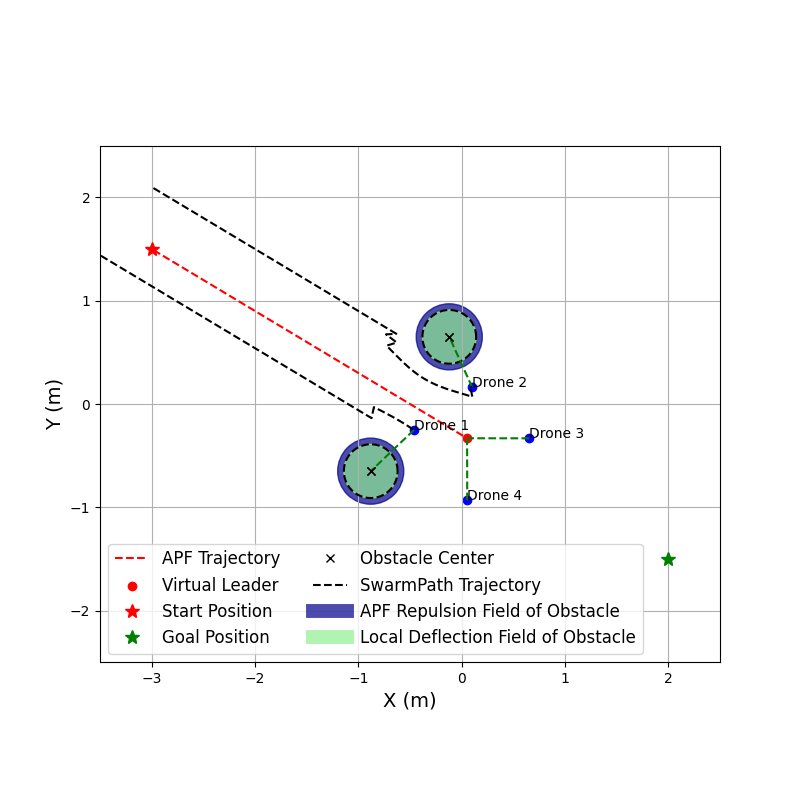} 
\vspace{-1.2cm}
\caption{Instance of simulation is shown in which drone 3 and drone 4 maintain leader-drone links whereas drone 1 and drone 2 temporarily breaks them in order to avoid collision from gate poles or obstacles. Blue circles of radius $r_{apf}$ represents the APF repulsion field, whereas green circles of radius $r_{imp}$ represents the region where the drone creates a link with an obstacle for deflection. The aim of the drones is to avoid gate poles via adaptive impedance links and reach final position by shortest distance.}
\label{sim_gates_only} 
\vspace{-0.3cm}
\end{figure}

In our setup, the APF trajectory does not directly influence the trajectory of drones in the local planner. Therefore, to address the issue of collision avoidance in local path planning, we allowed the drones to break their link with the APF trajectory while simultaneously creating a link with a nearby obstacle as shown in Fig.~\ref{sim_gates_only}, thus pushing the drone away from it. This dynamic link set-up enables the drone to effectively avoid the obstacle and also recreate its link with the APF trajectory once the obstacle is avoided. The deflection of the drones is calculated using equation \eqref{eq:deflection_eqn}.

\begin{equation}
\Delta x_{drone,n} = k_{impF} \cdot r_{imp}
\label{eq:deflection_eqn}
\end{equation}
where $r_{imp}$ is the radius of the local deflection region around the obstacle, $k_{impF}$ is the force coefficient adjusted to the drone's average velocity, and $n$ is the drone number. 

\subsubsection{{Analyzing Impedance Parameters}
\label{section:Impedance}}
For our experimental setup, impedance parameters, including mass of virtual link, spring constant, and damping coefficient were taken from \cite{Dandelion}. Parameters were selected to satisfy the critically damped response. In this section, we study the impact of impedance parameters on the overall trajectory of drones.

Table~\ref{Impedance_Parameters1} demonstrates the effect of varying damping coefficient by keeping the other parameters constant. It becomes clear how significantly the trajectory of a drone alters when the damping coefficient slightly surpasses its threshold. Furthermore, an increase in damping coefficient leads to an over-damped system resulting in a slow and sluggish system response. Given that these virtual impedance links connect drones and the virtual leader, which moves at its own pace, such adjustments can induced unexpected deviations in a drone's path. 
\begin{table}[h!]
    \centering
    \caption{Analyzing the Effect of Damping Coefficient, $N\cdot s/m$}
    \renewcommand{\arraystretch}{1} 
    \begin{tabular}{|c|c|c|c|c|} 
        \hline
        \multicolumn{5}{|c|}{\textbf{m = 1.9 $kg$, k = 20.88 $N/m$}} \\ 
        \hline
        \textbf{Trajectory (m) of:} & \textbf{d = 12.5} & \textbf{d = 12.6} & \textbf{d = 12.7} & \textbf{d = 12.8} \\
        \hline
        Drone 1 & 8.45 & 8.49 & 8.61 & 28.38 \\
        \hline
        Drone 2 & 8.46 & 8.49 & 8.64 & 29.68 \\
        \hline
        Drone 3 & 8.58 & 8.58 & 8.76 & 31.12 \\
        \hline
        Drone 4 & 8.63 & 8.63 & 8.77 & 29.05 \\
        \hline
    \end{tabular}
    \label{Impedance_Parameters1}
\end{table}

Subsequently, Table ~\ref{Impedance_Parameters2} illustrates the drone trajectory with different spring constant. Marginal effect is seen in drone trajectories if spring constant increases to 27 $N/m$, value exceeding this threshold exert the most pronounced influence. This is because increasing $K$ increases the natural frequency $\omega_n$ of oscillation of a system under the presence of external disturbance, which, if not counteract by the enough damping, can affect the system's stability leading to instability or uncontrolled oscillations. 
\begin{table}[h!]
    \centering
        \caption{Analyzing the Effect of Spring Constant,  $N/m$}
    \renewcommand{\arraystretch}{1} 
    \begin{tabular}{|c|c|c|c|c|} 
        \hline
        \multicolumn{5}{|c|}{\textbf{m = 1.9 $kg$, d = 12.6 $N\cdot s/m$}} \\ 
        \hline
        \textbf{Trajectory (m) of:} & \textbf{k = 20.8} & \textbf{k = 21} & \textbf{k = 28.0} & \textbf{k = 29.0} \\
        \hline
        Drone 1 & 8.45 & 8.49 & 8.66 & 9.75 \\
        \hline
        Drone 2 & 8.53 & 8.49 & 8.70 & 9.84 \\
        \hline
        Drone 3 & 8.66 & 8.58 & 8.84 & 10.64 \\
        \hline
        Drone 4 & 8.65 & 8.63 & 8.92 & 10.10\\
        \hline
    \end{tabular}
    \label{Impedance_Parameters2}
\end{table}

Table ~\ref{Impedance_Parameters3} demonstrates the impact of virtual mass value on the system; decreasing mass can make the system more responsive to the external forces thus accelerates more, resulting in larger trajectory. Likewise, spring constant, decreasing mass can also result in higher natural frequency $\omega_n$ of oscillations, so if damping is not enough, it can result in an unstable system.

\begin{table}[htbp]
    \centering
        \caption{Analyzing the Effect of Mass,  $kg$}
    \renewcommand{\arraystretch}{1.0} 
    \begin{tabular}{|c|c|c|c|c|} 
        \hline
        \multicolumn{5}{|c|}{\textbf{k = 20.88 $N/m$, d = 12.6 $N\cdot s/m$}} \\ 
        \hline
        \textbf{Trajectory (m) of:} & \textbf{m = 1.8} & \textbf{m = 1.9} & \textbf{m = 2.0} & \textbf{m = 2.1} \\
        \hline
        Drone 1 & 27.33 & 8.49 & 8.45 & 8.45 \\
        \hline
        Drone 2 & 28.73 & 8.49 & 8.45 & 8.45 \\
        \hline
        Drone 3 & 29.87 & 8.58 & 8.56 & 8.55 \\
        \hline
        Drone 4 & 27.87 & 8.63 & 8.64 & 8.63 \\
        \hline
    \end{tabular}
    \label{Impedance_Parameters3}
\end{table}

\subsection{Global Path Planning with APF}
\label{section:APF}
To avoid the obstacles and guiding drones from starting point to the destination, we applied the artificial potential field algorithm \cite{Improved_APF}. In this method, two virtual forces are acting. As the environment changes, this force field adapts accordingly, allowing the drones to move in a real-time environment. Equation of force field can be calculated as: 
\begin{equation}
F_{\text{total}} = F_{\text{attraction}} + F_{\text{repulsion}} \label{eq:total_force}
\end{equation}
where
\begin{align*}
F_{\text{attraction}}(d_{\text{g}}) &= k_{\text{att}} \cdot d_{\text{g}}, \\
F_{\text{repulsion}}(d_{\text{o}}) &= 
\begin{cases} 
0 & \text{if } d_{\text{o}} > d_{\text{safe}} \\
k_{\text{rep}} \cdot \left( \frac{1}{d_{\text{o}}} - \frac{1}{d_{\text{safe}}} \right) & \text{if } d_{\text{o}} \leq d_{\text{safe}}
\end{cases}
\end{align*}
where $d_{\text{g}}$ and $d_{\text{o}}$ are the distances from drone to goal and to obstacle, respectively, $k_{\text{att}}$ and $k_{\text{rep}}$ are the attraction and repulsion coefficients, respectively.

\begin{figure}[h!]
\centering
\includegraphics[width=0.98\linewidth]{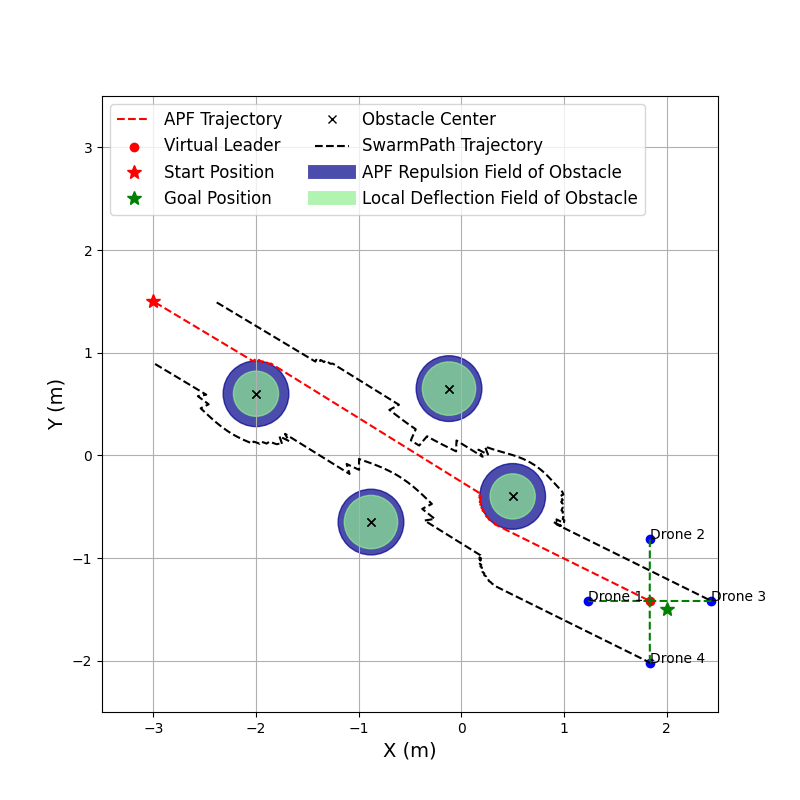} 
\caption{Fully executed SwarmPath trajectory involving four objects; two obstacles and two gate poles (placed in the middle of the obstacles). Drone 3 and drone 4 trails are drawn to show the navigation of agents in challenging scenarios.}
\label{sim_gates_obs}
\vspace{-0.35cm}
\end{figure}

The working of APF as a global path planner is shown in Python simulation when a virtual APF point reaches a goal position within a prescribed threshold.
Finally, the simulation was performed to test SwarmPath technology in a case where swarm agents avoid two obstacles placed near to starting and goal position and also avoid gate poles placed in the middle with slightly different radius than the obstacles as shown in Fig.~\ref{sim_gates_obs}. The trajectory shows the efficient execution of path planner that allows swarms to navigate through the shortest distance from starting to goal point using APF while maintaining the connection among the agents using adaptive impedance topology.

\subsection{Comparison with conventional APF}
\begin{figure}[h!]
    \centering
    \includegraphics[width=0.98\linewidth]{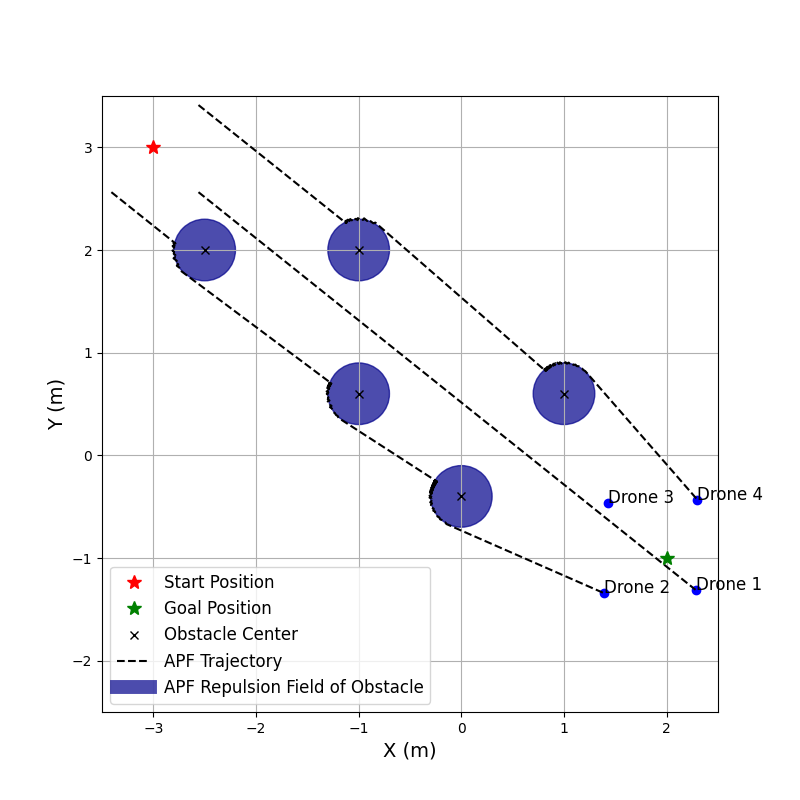}
    \caption{The trajectory generated by conventional APF showing potential connectivity losses among agents.}
    \label{fig:only APF}
\end{figure}

\begin{figure}[htbp]
    \vspace{-0.38cm}

    \centering
    \includegraphics[width=0.98\linewidth]{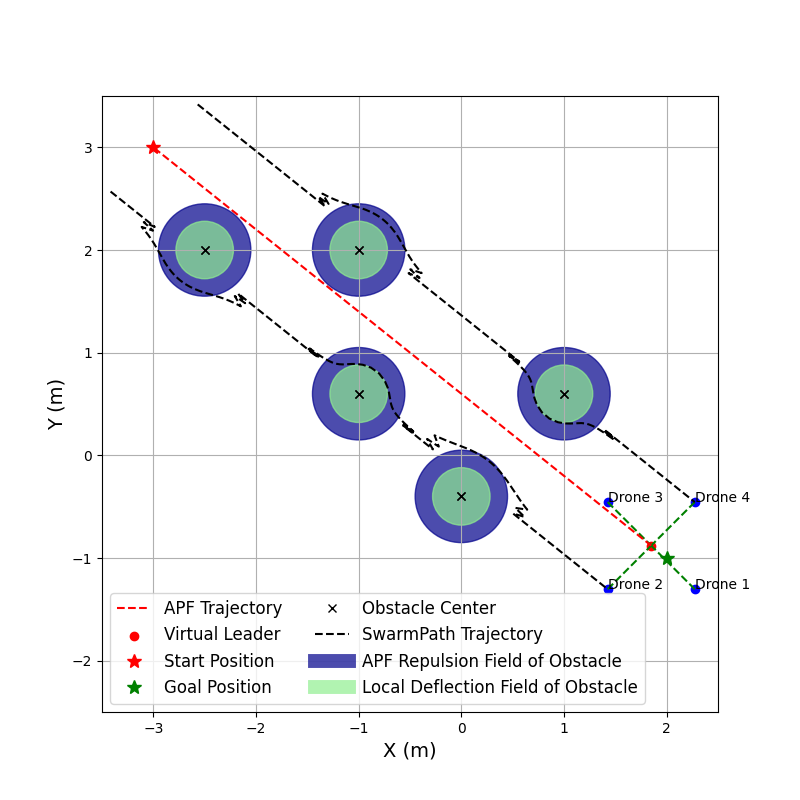}
    \caption{Integrated APF with mass spring damper system and impedance links presenting consistent connectivity between drones.}
    \label{fig:APF and impedance}
\vspace{-0.5cm}    
\end{figure}

A comparison between our technology and conventional APF was done to study the efficacy of SwarmPath approach by simulating more complex case of obstacles as in forests. The conventional APF generates a path for each drones to follow, but as shown in the Fig.~\ref{fig:only APF}, drone 2 and drone 4 navigates at the periphery of obstacles while drone 1 and drone 3 passes from the inside, ultimately disrupting the connectivity among swarm agents. In Table~\ref{Comparsion APF_IMP}, it can be seen that the maximum distance between drone 1 and drone 2 is more than twice in case of conventional APF as compared to SwarmPath. This lack of coordination in case of solo APF increases the risk of collision and interference between drones. To overcome this connectivity issue in the Fig.~\ref{fig:APF and impedance}, the APF repulsion field is replaced by impedance deflection field, while enlarging APF repulsion region slightly to create a safe collision free path globally. As shown in Table~\ref{Comparsion APF_IMP} the drone's maximum distance lies between 1.1 to 1.45 m, hence minimizing connection loss and thereby, highlighting the importance of impedance in terms of safety. Furthermore, total trajectory completion time decreased by $30\%$ when SwarmPath was deployed on agents, hence demonstrating greater efficiency.

\begin{table}[htbp]
    \centering
    \caption{Comparison of Conventional APF and APF with Impedance}
    \renewcommand{\arraystretch}{0.9}
    \begin{tabular}{|c|c|c|}
        \hline
        \textbf{\parbox{4cm}{\centering Drone 2 maximum relative \\ distance (m) from}} & \textbf{\parbox{2cm}{\centering Conventional \\ APF}} & \textbf{\centering SwarmPath} \\
        \hline
        Drone 1 & 2.75 & 1.21 \\
        \hline
        Drone 3 & 1.74 & 1.10 \\
        \hline
        Drone 4 & 2.31 & 1.45 \\
        \hline
    \end{tabular}
    \label{Comparsion APF_IMP}
\end{table}

\vspace{-12pt}

\section{Experimental Results}
\label{subsection:experiment}
\subsection{Experimental Setup}
Fig.~\ref{exp_setup_1} shows the experimental setup for testing of the algorithm. Path planning algorithm uses ROS2 nodes to publish position to the Crazyflie drones through the Crazyswarm \cite{crazyswarm} framework. The real-time positions of drones are captured with VICON motion capture system. The drones use an integrated Crazyflie PID controller for trajectory following \cite{CF_PID}. Therefore, drones dynamics and controllers were developed in simulation environment for better reflection of the real environment.

\begin{figure}[h!]
\centering
\vspace{0.2cm}
\includegraphics[width=0.90\linewidth]{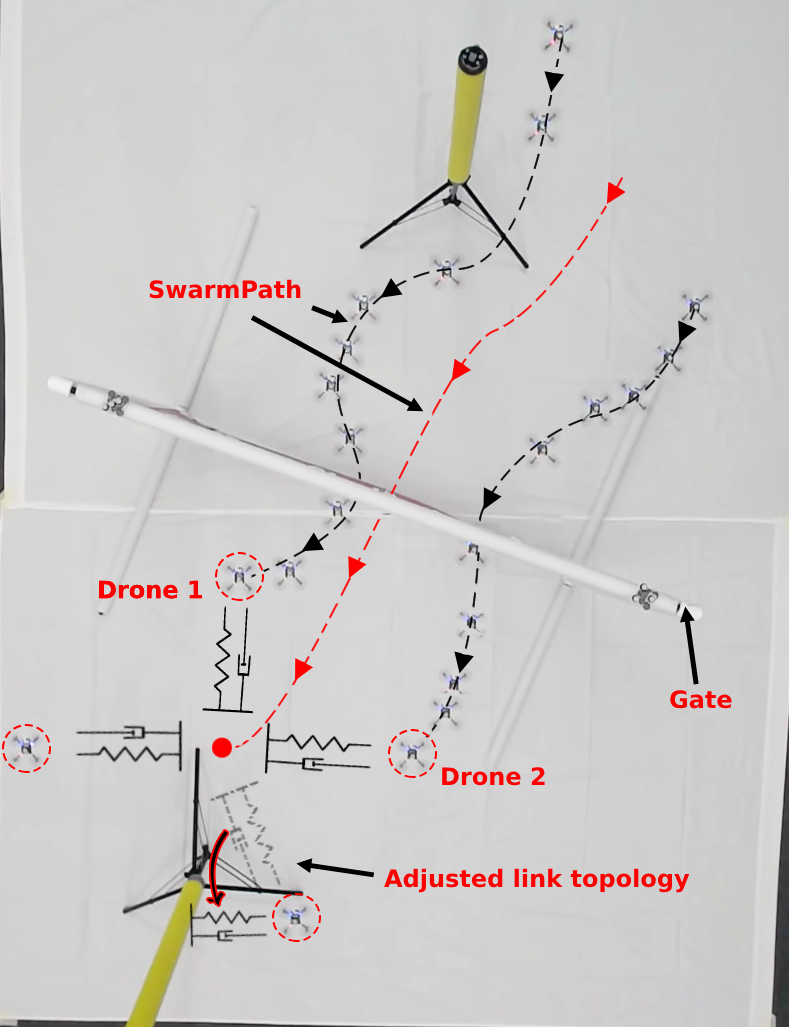} 
\caption{Swarm agents navigating through gate and obstacles in real-world. Trajectories of only drone 1 and drone 2 are drawn for clarity.}
\label{case_a_top_view}
\vspace{-0.5cm}

\end{figure}

\subsection{Results} 
\begin{figure}[htbp]
\centering

\includegraphics[width=0.98\linewidth]{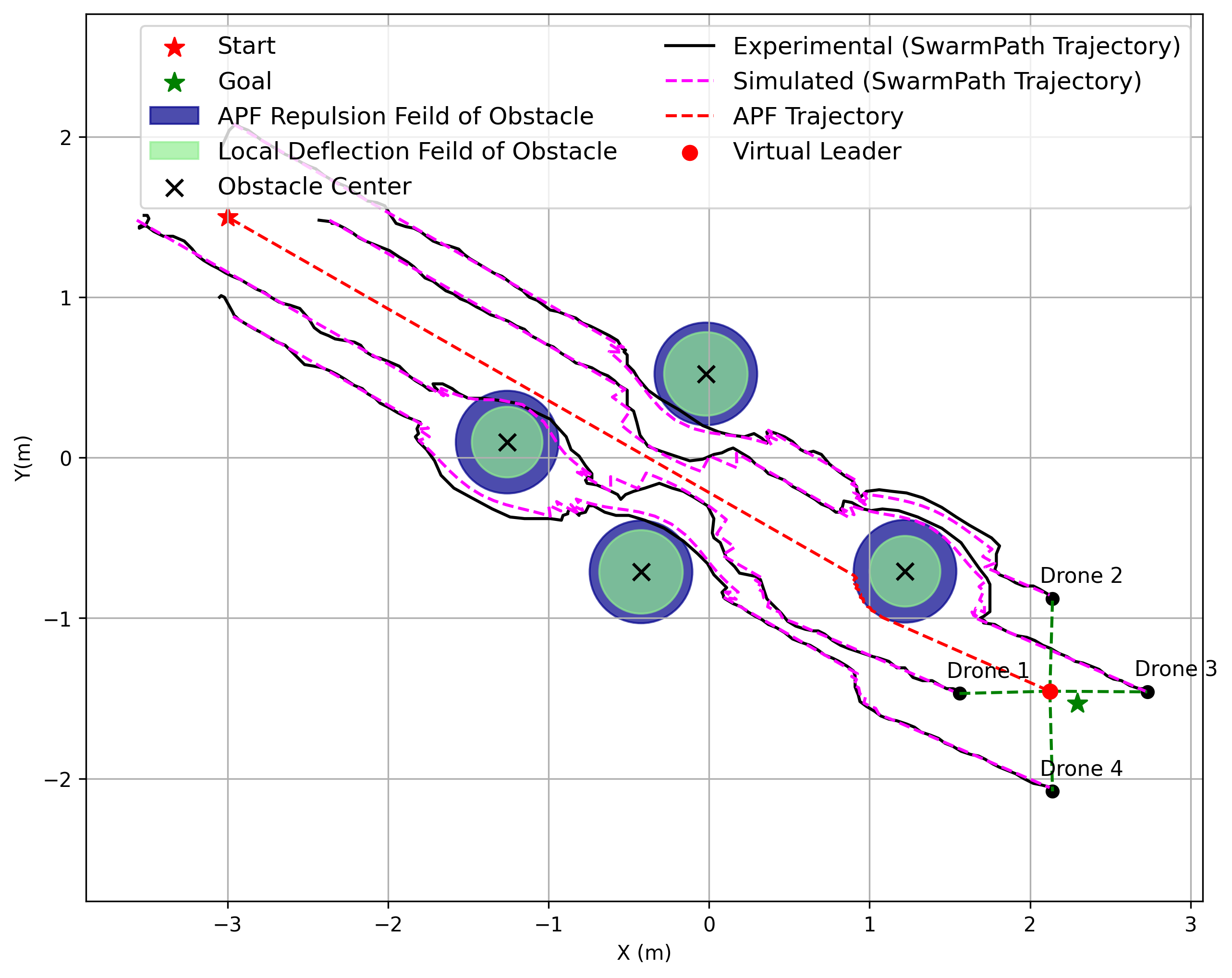} 
\caption{Comparison of swarm agents navigating through gate and obstacles in real-world and simulation.}
\label{case_1_comparison}

\end{figure}

\begin{figure}[h!]
\centering
\includegraphics[width=0.98\linewidth]{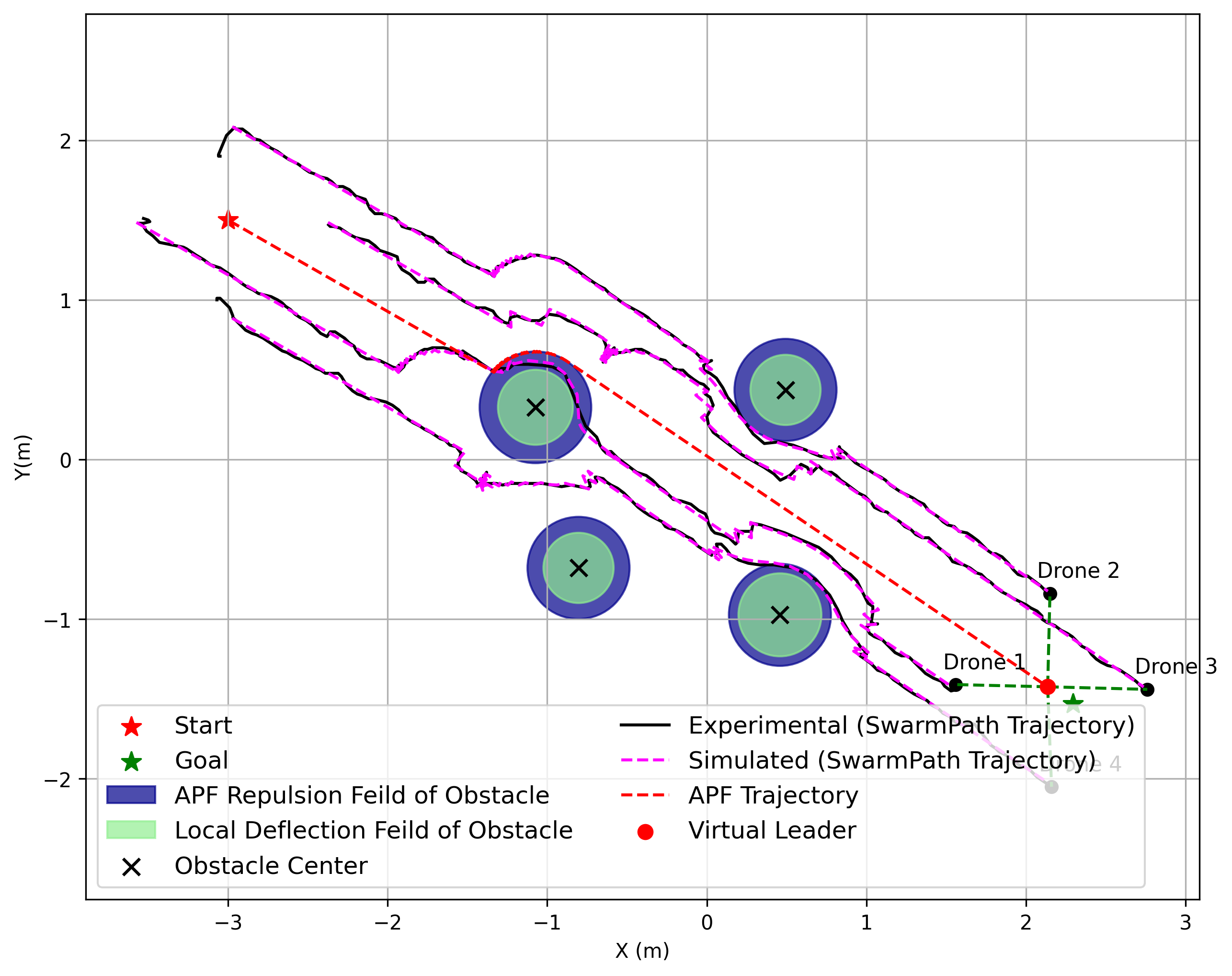} 
\caption{Comparison of swarm agents navigating through complex obstacle terrain where impedance link dampen the oscillatory motion caused by APF local minima at first obstacle.}
\label{case_2_comparison}
\end{figure}

The experiment was performed on two cases. In Case 1, as shown in Fig.~\ref{case_1_comparison}, the objective of drones was to avoid two obstacles while also pass through the center of gate as shown in Fig.~\ref{case_a_top_view}. In Case 2, Fig.~\ref{case_2_comparison}, the gate was replaced by two more obstacles, therefore, testing the capability of algorithm to avoid all the obstacles, in forest-like environment.

\subsubsection{Error between Experiments and Simulations}
Table~\ref{Error} shows the comparison of each drone trajectory at all time stamps in simulated and real environment. The results show that the total absolute percentage error (APE) of trajectory of all drones in Case 1 is approximately 1\% smaller than in Case 2, indicating that the algorithm performs slightly better when navigating through gates rather than avoiding multiple convex obstacles. In both cases, the mean error did not exceed 10\% that proves the implementation of algorithm is practical from simulation to reality. The APE value can be improved in future by tuning the control parameters of drones. 

\begin{table}[htbp]
    \centering
    \caption{Error between Simulated and Experimental Trajectories for Each Drone}
    \renewcommand{\arraystretch}{1}
    \begin{tabular}{|c|c|c|} 
        \hline
        \multicolumn{3}{|c|}{\textbf{Error}} \\ 
        \hline
        \textbf{Drone} & \textbf{Case I} & \textbf{Case II} \\
        \hline
        1 & 5.24\% & 7.49\% \\
        \hline
        2 & 6.36\% & 3.00\% \\
        \hline
        3 & 6.01\% & 8.55\% \\
        \hline
        4 & 7.60\% & 10.00\% \\
        \hline
       \textbf{Total trajectory error} & 6.31\% & 7.37\% \\ 
        \hline
    \end{tabular}
    \label{Error}
\end{table}

Table~\ref{trajectory} indicates that in Case 2, drones cover a greater distance in the real environment compared to the simulated one. The discrepancy is expected because in real-world conditions, drones encounter the real complex obstacle environment, hence, leading to its more oscillatory behavior, which also supports it higher APE value in Table~\ref{Error}. 
\begin{table}[h!]
    \centering
    \caption{Swarm Trajectory Length in Real and Simulated Environments for Two Experimental Cases}
    \renewcommand{\arraystretch}{1}
    \begin{tabular}{|c|c|c|c|c|} 
        \hline
        \multicolumn{5}{|c|}{\textbf{Trajectory Length (m)}} \\ 
        \hline
        \textbf{Drone} & \multicolumn{2}{c|}{\textbf{Real Environment}} & \multicolumn{2}{c|}{\textbf{Simulated Environment}} \\
        \hline
        & Case 1 & Case 2 & Case 1 & Case 2 \\
        \hline
        1 & 6.70\ & 7.45\ & 7.07\ & 8.05 \\
        \hline
        2 & 6.69\ & 6.78\ & 7.15\ & 6.99 \\
        \hline
        3 & 6.67\ & 6.86\ & 7.09\ & 7.50 \\
        \hline
        4 & 6.62\ & 7.07\ & 7.17\ & 7.86 \\
        \hline
    \end{tabular}
    \label{trajectory}
\end{table}

\section{Limitations}

Although the proposed solution provides implementable results with an easy-to-go algorithm, a number of drawbacks must be recognized to fully comprehend its applicability. 

While performing experiments, VICON was used for detecting obstacles and positions of drones, therefore, the presence of varying lightning conditions can have substantial effects on the performance of the algorithm. 
 
In addition, the algorithm is tested with the assumption that the location of obstacles is known. Therefore, this opens a wide area in the field of Robotics to apply simultaneous localization and mapping (SLAM) techniques to create swarms of drones fully autonomous.

\section{Conclusion and Future work}

In this study, the proposed SwarmPath system is a novel and easy-to-go algorithm for dynamic interaction between APF and adaptive impedance control by creating virtual links of drones with the obstacles for collision avoidance as a local path planner. Our experimental results demonstrated the efficacy of the algorithm in real-world scenarios. The purpose of this study was to create a simple yet effective solution that can be implemented on different and complex environmental conditions.

Through two different environment cases, we evaluated that the algorithm performed proficiently but slighltly better in the case of flying through the gate as compared to maneuvering around complex obstacles arrangement by comparing their respective APE values with simulation results. This is because, in Case 2, obstacles were densely packed creating a cluttered environment and making it difficult for drones to navigate through them as in forest. Nonetheless, the APE value of drone trajectories in two different approaches was about $6-7\%$. The APE can be further improved by tuning the controller parameters, promising even greater performance in future iterations. Moreover, by adopting the SwarmPath approach, the drone's trajectory completion time was reduced by $30\%$ when compared to conventional APF. Additionally, the maximum distance between drone 1 and drone 2 decreased by $56\%$ , demonstrating enhanced efficiency and improved connectivity between the drones.
    
Overall, our findings highlight the potential of integrating two different approaches that enable safe and efficient navigation of swarms of drones. By addressing the limitations and refinements of algorithms such as incorporating SLAM, the applicability of our algorithm can be extended to various domains by creating system of drones fully autonomous. In forest environments, drones could navigate through passages to track wildlife and vegetation and gather valuable data contributing to ecosystem management. Moreover, the versatility of the algorithm can extend to the realm of swarm racing. The user can apply this algorithm to pass through different gates, thus representing a novel and engaging application to revolutionize the field of drone racing.

\section*{Acknowledgements} 
Research reported in this publication was financially supported by the RSF grant No. 24-41-02039.

\bibliographystyle{IEEEtran}
\bibliography{bibliography}
\balance
\addtolength{\textheight}{-12cm}
\end{document}